\documentclass[11pt]{article}

\usepackage[preprint]{acl}

\usepackage{times}
\usepackage{latexsym}
\usepackage{xcolor} 
 \usepackage{amsfonts}
 \usepackage{booktabs}
\usepackage[T1]{fontenc}

\usepackage[utf8]{inputenc}

\usepackage{microtype}

\usepackage{inconsolata}

\usepackage{graphicx}
\usepackage{amsmath}
\title{Translating in style: An open-weight alternative to GPT-4 for culturally aligned machine translation}

\author{
Thomas Rietman$^{1,*}$ \
Hadi Mohammadi$^{1}$ \
Robert A. Bagheri$^{1}$\\[0.5em]
$^{1}$Department of Methodology and Statistics, Utrecht University, The Netherlands
}

\begin{document}
\maketitle
\begin{abstract}
Machine translation systems often preserve meaning but systematically flatten
linguistic style, a failure that is most acute when translating into
non-Western languages. The RASTA framework of \citet{havaldar2025atowards} addresses
this through retrieval-augmented stylistic alignment, but relies on GPT-4, a
closed, proprietary model. We investigate whether an open-weight translation
specialist, Hunyuan-MT-7B, can serve as an open-weight alternative. We evaluate
three variants of the model, the base model, a LoRA supervised fine-tune, and
a GRPO-steered policy. We do this across three style dimensions (formality, politeness,
and intimacy) and twelve directed language pairs, measuring style alignment
with the correlation-based metric $\mathcal{A}$. We find that fine-tuning does
not reliably improve alignment: the three variants are statistically
indistinguishable in aggregate. The open backbone does not match RASTA-GPT-4
overall, but the shortfall is concentrated in formality and largely closes for
intimacy and Chinese directions, where the open model matches or surpasses
GPT-4 on individual pairs. An accessible RASTA pipeline built on an open model
is thus already viable for parts of the problem.
\end{abstract}


\section{Introduction}
In our increasingly interconnected world, machine translation (MT) has become essential infrastructure for cross-cultural dialogue \citep{khasawneh2023potential, shadiev2016facilitating}. While Large Language Models (LLMs) are increasingly adopted to facilitate translation, recent studies show a systemic failure: these systems preserve semantic content but systematically fail to preserve style \citep{havaldar2025atowards, kajava2020emotion, troiano2020lost}.
Linguistic style refers to the systematic variation in linguistic choices across contexts and speakers combined with grammatical and lexical features that signal social identity, attitude, and communicative intent \citep{biber2019register}. Beyond just aesthetics, style is vital to how a language works: speakers deploy it to signal respect and manage social relationships for example. \citep{danescu2013computational, kang2020linguistically, kang2021stylesinglevariablecase}. How style is encoded differs substantially across languages and cultures. In Japanese, for instance, formality is deeply embedded in an elaborate system of honorifics and verb conjugations, while English encodes formality more explicitly and lexically \citep{feely2019formality}. A translation may thus be semantically accurate but lacking in formality: misalignment between intended and perceived style can cause confusion, misunderstanding or offense \citep{havaldar2025atowards, moorjani1988semiotic, hall1976beyond}.
One key factor underlying these stylistic inconsistencies may be the cultural bias embedded in LLMs. Existing models are strongly shaped by Western, Anglocentric training data and design choices, leading them to internalize Western communication norms and to perform systematically worse when capturing non-Western stylistic variation \citep{adilazuarda2024towards, johnson2022ghostmachineamericanaccent, dwivedi2023eticor, havaldar-etal-2023a-multilingual, atari2023whichhumans}. This imbalance is quantified by \citet{havaldar2025atowards}, who introduce a style alignment metric $\mathcal{A}$.
A measures the product-moment correlation between the style of a source text and its translation across three dimensions: politeness, intimacy, and formality. They evaluated the performance of different LLMs including GPT-4, Google Translate, and Llama-3.2. They find that style alignment is consistently worst when translating into non-Western languages such as Japanese and Chinese. Moreover, LLMs systematically bias translations toward stylistic neutrality, reducing the natural variance present in real-world communication and erasing culturally specific expression \citep{plum2025identity}.
To address these failures, \citet{havaldar2025atowards} introduce RASTA (Retrieval-Augmented STylistic Alignment), a method that learns cultural alignment mappings in embedding space and retrieves culturally appropriate few-shot exemplars to guide translation. Using GPT-4, RASTA achieves up to 56\% improvement in style alignment over vanilla translation baselines, without significant degradation in standard translation quality metrics, and is preferred by native bilingual speakers in human evaluation. However, RASTA's reliance on GPT-4 as the underlying generation model limits its accessibility its a closed source, paid model. Instead, we replace GPT-4 with Hunyuan-MT-7B, an open-source 7B-parameter multilingual translation model developed by Tencent. Hunyuan-MT-7B is specifically optimized for translation and supports 33 languages. Despite its relatively compact size, it achieves state-of-the-art performance among similarly sized models and ranked first in 30 out of 31 language pairs in the WMT 2025 General Machine Translation shared task. Its specialized training framework and open-source availability make it a practical and cost-effective alternative for multilingual translation \cite{hunyuanmt2025}.
This raises the question: \textit{Can a fine-tuned open-weight model, specifically Hunyuan-MT-7B, achieve comparable style alignment to RASTA-GPT-4 while improving accessibility and reproducibility?}
Building on the RASTA framework of \citet{havaldar2025atowards}, this study investigates whether targeted improvements on the translating model of the pipeline can advance style alignment. Concretely, we will look if we can get similar performance to GPT-4 by tuning a specialized translation model.


\section{Literature review}

\subsection{Style in Translation}
Successful translation requires more than the preservation of semantic content: it must also transfer the stylistic properties of the source text into the target language \citep{boase2004translation, nida1982theory}. Linguistic style encompasses systematic variation in vocabulary, grammar, and discourse structure that signals social identity, attitude, and communicative intent \citep{biber2019register}, and includes dimensions such as formality, politeness, and intimacy \citep{havaldar2025atowards}. These dimensions are not universal: they are culturally and linguistically encoded in different ways across languages, making their preservation in translation particularly challenging \citep{hershcovich2022challenges}.

An example of this challenge is the translation of formality between English and Japanese. \citet{feely2019formality} demonstrate that English-to-Japanese translation tends to lose formality, as Japanese encodes it through a rich system of honorifics and verb conjugations that have no direct English equivalent. Their rule-based and heuristic approaches targeting these specific grammatical structures show that formality preservation requires language-specific knowledge beyond what standard MT systems provide. More broadly, \citet{troiano2020lost} and \citet{kajava2020emotion} demonstrate that stylistic and emotional properties are frequently lost in MT, even when semantic content is preserved, highlighting that style transfer in translation remains challenging.

\subsection{Style-Controlled Machine Translation}
A substantial body of work has approached style preservation in MT through model fine-tuning. \citet{rippeth2022controlling} demonstrate that pre-trained multilingual language models outperform bilingual counterparts in preserving formality, though they observe a trade-off between stylistic preservation and translation fluency that remains difficult to resolve simultaneously. Similarly, \citet{tyagi2023} fine-tune an IndicBERT model on formality-annotated data and find that large multilingual models can outperform smaller bilingual models in style-controlled translation.

Beyond fine-tuning, prompting and retrieval-augmented strategies have emerged as promising alternatives. \citet{sarti-etal-2023-ramp} introduce RAMP, a framework that guides LLMs in style-preserving translation by prompting with retrieved stylistically similar sentences and explicitly marked style attributes. Their approach demonstrates that carefully constructed prompts can help LLMs retain stylistic information without requiring architectural changes. Building on this line of work, \citet{havaldar2025atowards} introduce RASTA (Retrieval-Augmented STylistic Alignment), which learns cultural alignment mappings in embedding space using a BGE-M3 model \citep{chen2024bge} and retrieves the most stylistically appropriate native-language exemplars as few-shot context during translation. Using GPT-4, RASTA achieves up to 56\% improvement in style alignment over vanilla translation baselines, without significant degradation in standard translation quality metrics, and is preferred by native bilingual speakers in human evaluation.

\subsection{Cultural Bias in LLMs}
Many widely used LLMs, are trained on Western, English data. \citet{adilazuarda2024towards} provide a systematic survey of how culture is represented and modelled in LLMs, finding consistent underrepresentation of non-Western perspectives. \citet{dwivedi2023eticor} further demonstrate that LLMs fail to capture culturally specific etiquette norms outside the Anglophone sphere. In the context of multilingual modelling, \citet{havaldar-etal-2023a-multilingual} show that multilingual LLMs are not multicultural: despite processing multiple languages, these models internalize predominantly Western cultural norms, leading to worse performance in culturally sensitive tasks such as emotion and style recognition.

\subsection{Multilingual and Open-Weight Models}
Recent years have seen substantial progress in open-weight, instruction-tuned LLMs that approach the capabilities of proprietary systems on a range of NLP tasks. Mistral-7B \citep{jiang2023mistral} achieves strong performance on reasoning and language understanding benchmarks despite its relatively compact size. Llama-3 \citep{dubey2024llama} and Gemma \citep{gemmateam2024gemma} similarly demonstrate competitive multilingual capabilities. However, \citet{zhao2024llama} show that open-weight models such as Llama still exhibit significant performance gaps on non-English tasks, and that multilingual capability does not straightforwardly transfer across languages without targeted adaptation. Large-scale multilingual models such as BLOOM \citep{workshop2023bloom176bparameteropenaccessmultilingual} have been developed explicitly to address language diversity, but coverage of culturally specific stylistic phenomena remains limited.

In the context of MT specifically, \citet{havaldar2025atowards} evaluate Llama-3.2-11B within the RASTA framework and observe that, unlike GPT-4, Llama is overly reliant on few-shot exemplars, sometimes incorporating information from multiple exemplars into a single translation, leading to degradation in translation quality as measured by GEMBA and CometKiwi. This suggests that the interaction between model architecture, instruction-following capability, and retrieval-augmented prompting is affecting the models, and that open-weight models may require model-specific fine-tuning or prompt adaptation to fully benefit from stylistic retrieval frameworks.

\subsection{Evaluation of Style in Translation}
Standard MT evaluation metrics such as BLEU \citep{papineni2002bleu} and BERTScore \citep{zhang2020bertscore} are designed to capture fluency and semantic adequacy, but are not sensitive to stylistic dimensions \citep{havaldar2025atowards, mukherjee2025}. More recent reference-free metrics, including GEMBA \citep{kocmi-federmann-2023-large} and CometKiwi \citep{rei-etal-2022-cometkiwi}, improve on surface-level matching but still show low correlation with style alignment as measured by the metric $\mathcal{A}$ introduced by \citet{havaldar2025atowards}. \citet{mukherjee2025} conduct a comprehensive comparison of existing evaluation metrics and propose hybrid alternatives, finding that while their hybrid metrics outperform individual baselines, none surpasses GPT-4-based prompted evaluation for style assessment, further emphasising the strength of large-scale LLMs as style evaluators but also the absence of a reliable automated style-sensitive metric.

To address this gap, \citet{havaldar2025atowards} train a set of 14 Mistral-7B regression models, one per language and style dimension, as style quantifiers, achieving low RMSE on held-out test sets. These quantifiers form the backbone of the $\mathcal{A}$ metric.


\section{Methods}

\subsection{The RASTA Framework}
This study builds upon the RASTA framework introduced by \citet{havaldar2025atowards}. 
We briefly summarise the original pipeline before describing proposed modifications.

\subsubsection{Style Alignment Metric}\label{3.1.1}
\citet{havaldar2025atowards} introduce a style alignment metric $\mathcal{A}$ to 
quantify the degree to which the intended style of a source text is preserved in its 
translation. Given two language-specific style quantifiers $\mathcal{C}_1 : L_1 
\rightarrow [0,1]$ and $\mathcal{C}_2 : L_2 \rightarrow [0,1]$, style alignment is 
defined as the product-moment correlation between the style scores of the original 
and translated texts across a corpus $X$:

\begin{equation}
\mathcal{A}(L_1, L_2) = r\left(\mathcal{C}_1(X_{L_1}),\ 
\mathcal{C}_2\left(T(X_{L_1})\right)\right)
\end{equation}

\noindent where $T : L_1 \rightarrow L_2$ is a translation function. This metric 
is computed across three style dimensions: politeness, intimacy, and formality, 
using 14 Mistral-7B regression models fine-tuned separately for each language and 
style dimension on multilingual corpora annotated by native speakers 
\citep{havaldar-etal-2023b-comparing, pei-etal-2023-semeval, rao-tetreault-2018-dear, 
briakou-etal-2021-ola}. These style quantifiers achieve a mean test RMSE of 0.157, 
0.183, and 0.255 for politeness, intimacy, and formality respectively.

\subsubsection{Retrieval-Augmented Stylistic Alignment}\label{rasta}
To improve style preservation during translation, \citet{havaldar2025atowards} 
introduce the RASTA method. RASTA operates in embedding space using a BGE-M3 
multilingual embedding model \citep{chen2024bge}. For every source language $L_1$ 
and target language $L_2$, RASTA computes two vectors: a \textit{native mapping} 
$\mathbf{v}_\text{native}$, which captures the shift between stylistically equivalent 
text in $L_1$ and $L_2$ as produced by native speakers, and a \textit{translation 
mapping} $\mathbf{v}_\text{trans}$, which captures the shift between text in $L_1$ 
and its machine translation into $L_2$. A cultural alignment vector is then derived as:

\begin{equation}
\mathbf{v}_\text{align} := \mathbf{v}_\text{native}(L_1, L_2, S) - 
\mathbf{v}_\text{trans}(L_1, L_2, S)
\end{equation}

\noindent This alignment vector is applied to the embedding of a new source text, 
transforming it to lie closer to where a native speaker of $L_2$ would position 
text of the same style $S$. The five most similar native $L_2$ sentences are then 
retrieved by cosine similarity and used as few-shot exemplars in the translation 
prompt. In the original work, GPT-4 serves as the underlying translation model, 
achieving up to 56\% improvement in $\mathcal{A}$ over a vanilla translation 
baseline with less than 1.5\% degradation in CometKiwi and GEMBA scores.

\subsubsection{Language Pairs}
Since this study uses the same dataset as \citet{havaldar2025atowards}, we are constrained to the language pairs available in their work. Due to time and computational resource limitations, we select a subset of these language pairs for our experiments. The selected language pairs differ across style dimensions and are listed below:

\begin{itemize}
\item \textbf{Intimacy:} Spanish--French (Es--Fr), English--Chinese (En--Zh)
\item \textbf{Politeness:} Spanish--Japanese (Es--Ja), English--Chinese (En--Zh)
\item \textbf{Formality:} English--French (En--Fr), Italian-Portuguese (It--Pt)
\end{itemize}

For each style dimension, we intentionally selected one language pair involving English and one pair without English. This gives us a better overview of the models potential to be used for different languages.

\subsection{Fine-tuning-Hunyuan-MT-7B for Stylistic Translation}

While RASTA substantially improves style alignment, its reliance on GPT-4 as the 
generation model introduces practical limitations: GPT-4 is a closed-source, 
proprietary model with associated cost and reproducibility constraints, and as 
\citet{havaldar2025atowards} note,  Llama-3.2-11B struggled within the RASTA framework due to over-reliance on few-shot exemplars. This study therefore investigates whether a smaller, open-weight model can be 
fine-tuned to serve as an effective drop-in replacement for GPT-4 in the RASTA pipeline.

We select Hunyuan-MT-7B \citep{hunyuanmt2025} as our base model. Hunyuan-MT-7B is specifically optimized for translation and supports 33 languages.

\subsubsection{Stage 1: Supervised Fine-tuning on Parallel Data}\label{stage 1}
The first stage of adaptation establishes general translation competence in a prompt format that the model can later be steered with. The prompt can be found in the appendix~\ref{prompt} We fine-tune Hunyuan-MT-7B using Low-Rank Adaptation \citep{hu2022lora}, training low-rank adapters on the attention and feed-forward projections of the base model in bfloat16 precision. The base weights are frozen throughout, so only the adapter parameters change. Important paramters can be found in Table~\ref{tab:stage1-hparams}
Training data is drawn from the OPUS collection \citep{4992de1b5fb34f3e9691772606b36edf} for the relevant language pair. Only human-made subsets were used. Exact data usage for each language pair can be found in the appendix~\ref{sec:dataused}. 
We format the prompt used during this stage to be as much like the RASTA prompt that will be used at the end during inference. We adopt a structured prompt inspired by the "preserve style" template of \citet{havaldar2025atowards}, stripped of the five retrieved exemplars and the explicit style-level line: the model is told that style and tone vary across cultures, given the source text, and asked to translate. The exact prompt is in the appendix~\ref{prompt}. Each parallel sentence pair contributes two training examples, one for each translation direction. Examples are wrapped using the model's native chat template, with loss computed only on the target translation. 
\begin{figure*}[t]
    \centering
    \hspace*{-2.4cm}
    \includegraphics[width=1.28\textwidth]{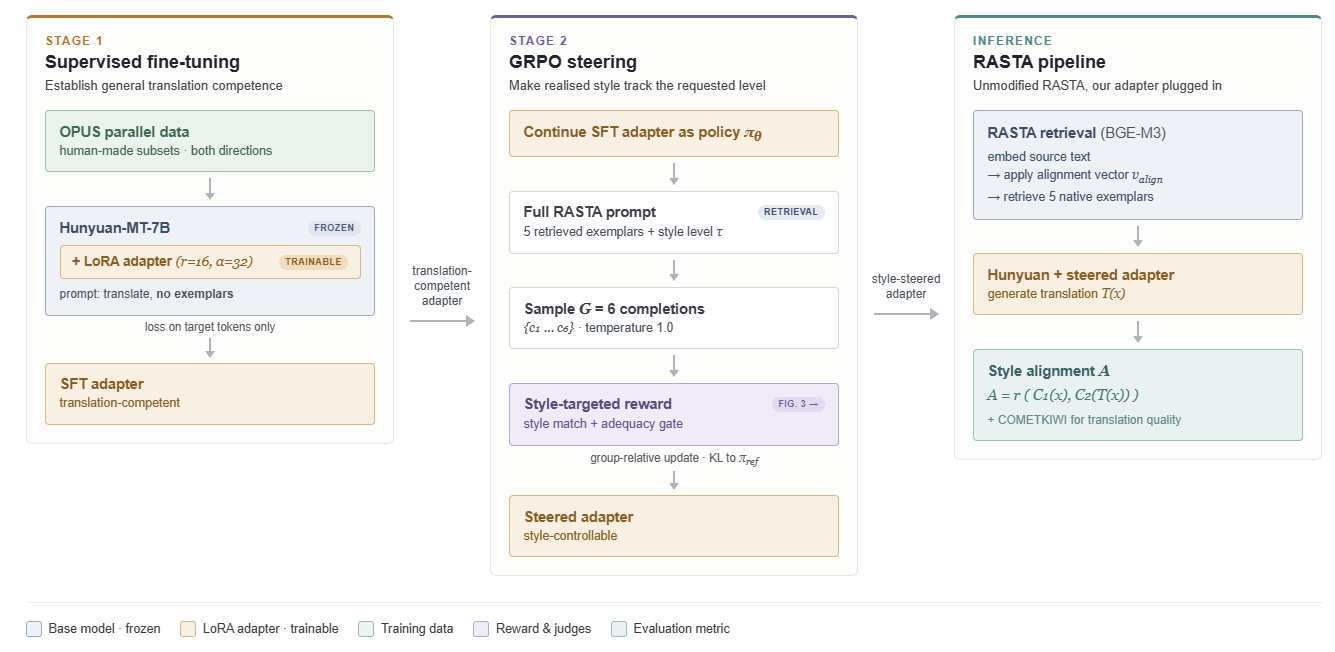}
    \caption{Overview of all the stages in our pipeline}
    \label{fig:step-1}
\end{figure*}

\begin{table}[ht]
\centering
\small
\caption{Stage 1 supervised fine-tuning configuration. Low-rank adapters are trained on Hunyuan-MT-7B with the base weights frozen. The same configuration is used for every (language pair, style dimension) setting.}
\label{tab:stage1-hparams}
\begin{tabular}{ll}
\hline
\textbf{Hyperparameter} & \textbf{Value} \\
\hline
\multicolumn{2}{l}{\textit{LoRA adapter}} \\
\quad Rank ($r$)              & 16 \\
\quad Scaling ($\alpha$)      & 32 \\
\quad Dropout                 & 0.05 \\
\quad Target modules          & \texttt{q\_, k\_, v\_, o\_,} \\
                              & \texttt{gate\_, up\_, down\_proj} \\
\hline
\multicolumn{2}{l}{\textit{Optimisation}} \\
\quad Optimiser               & AdamW \\
\quad Learning rate           & $2\times10^{-4}$ \\
\quad LR schedule             & Cosine \\
\quad Warmup ratio            & 0.05 \\
\quad Weight decay            & 0.01 \\
\hline
\multicolumn{2}{l}{\textit{Training schedule}} \\
\quad Epochs                  & 3 \\
\quad Per-device batch size   & 16 \\
\quad Gradient accumulation   & 1 \\
\quad Effective batch size    & 16 \\
\hline
\multicolumn{2}{l}{\textit{Precision \& memory}} \\
\quad Adapter precision       & bfloat16 \\
\quad Gradient checkpointing  & Enabled \\
\hline
\end{tabular}
\end{table}

\subsubsection{Stage 2: Steering with Group Relative Policy Optimisation}
\label{stage 2}
 
The supervised stage of Section~\ref{stage 1} attempts to improve translation competence,
but it does not make the model responsive to a
requested style level: the target level appears in the prompt, yet the model is free
to ignore it. The second stage therefore optimises the adapter so that the measured
style of the generated translation tracks the requested target, using reinforcement learning with a style-targeted reward.
 
We first attempted to induce this behaviour with Direct Preference Optimisation
\citep{NEURIPS2023_a85b405e}, constructing preference pairs from candidate translations
generated across a grid of target levels. This proved ineffective: the within-group
style contrast was too small to yield usable pairs, and the resulting policy showed essentially no
steerability (style--target correlation $\approx 0.05$).
Because DPO learns only from discrete pairwise comparisons, it has no signal to exploit
when sampled candidates differ only marginally in style. We therefore adopt Group
Relative Policy Optimisation (GRPO) \citep{shao2024deepseekmath}, which derives a
gradient from the relative reward of several completions sampled for the same
prompt. Even small reward differences within a group produce a learning signal, and
these differences compound over many online updates, which we found necessary to
get the steerability that single-shot sampling could not surface.
 
\paragraph{Training prompts.}
Unlike Stage~1, the GRPO prompts reinstate the full RASTA template: the five
retrieved native exemplars and the explicit style-level line are restored, so that the
model is trained on exactly the prompt it will receive at inference. For each source
sentence we construct the cultural alignment vector $\mathbf{v}_\text{align}$ and
retrieve exemplars as in Section~\ref{rasta}, and we instantiate the prompt at the whole number style levels per data set.
Each (source sentence, target level) pair forms one training row carrying the source text
$x$ and a normalised target level. 
 
\paragraph{Reward.}
The reward has two components: a style reward and an adequacy gate that rejects translations that deviate from the source meaning. To measure style, we use the target-language style quantifier $\mathcal{C}_2$ introduced in Section~\ref{3.1.1}, implemented as a frozen Mistral-7B regression head. Because the head's outputs are not naturally bounded, we linearly calibrate them to the $[0,1]$ style scale using 1,000 target-language sentences from the training corpus. The calibration is fitted by least-squares regression between the model outputs and the normalised style labels. The resulting predictions correlate strongly with the annotated style labels ($r = 0.947$), indicating that the quantifier provides a reliable style signal for reward computation. We measure adequacy as the BGE-M3 cosine similarity $\mathrm{sim}(x,c)$ between
the source $x$ and a candidate translation $c$. Given a target $\tau$, the reward is
\begin{equation}
R(c \mid x, \tau) =
\begin{cases}
-\,\bigl|\,\hat{\mathcal{C}}_2(c) - \tau\,\bigr| & \text{if } \mathrm{sim}(x,c) \ge \delta,\\[4pt]
\phi & \text{otherwise,}
\end{cases}
\end{equation}
with adequacy threshold $\delta = 0.5$ and floor $\phi = -2.0$. When a translation is
on-topic, its reward is highest (zero) when its style matches the target. Since both
$\hat{\mathcal{C}}_2(c)$ and $\tau$ fall in $[0,1]$, the lowest possible on-topic reward
is $-1$. We set the floor below this ($\phi < -1$) so that an off-topic translation can
never score better than an on-topic one in the same group, which stops the model from
improving its style score by abandoning the meaning. The style head and the embedding model are frozen and act only as judges. We note
that $\hat{\mathcal{C}}_2$ belongs to the same family of Mistral-7B style regression
heads used to compute the evaluation metric $\mathcal{A}$ in
Section~\ref{3.1.1}: both are trained on the same annotated corpora for the same
style dimension, and differ only in that the reward head is additionally calibrated
(via least-squares or Platt scaling, Section~\ref{stage 1}) to produce a dense
scalar suitable for reward shaping. The reward signal used to train GRPO is
therefore not independent of the metric used to evaluate it; we return to this
point when interpreting GRPO's results in Section~\ref{sec:results-finetuning}.
 
\paragraph{Objective.}
For each prompt, GRPO samples a group of $G = 6$ completions from the current model
$\pi_\theta$ and scores them with the reward. Each completion's advantage is its
reward relative to the group: the group's mean reward is subtracted off (and the result
scaled by the group's standard deviation), so a completion is reinforced when its reward
is above the group average and suppressed when it is below. The model is then updated to
increase the probability of the above-average completions, with two safeguards: the
per-token update is clipped so no single step moves the policy too far, and a KL penalty
keeps it close to a fixed reference model $\pi_{\text{ref}}$,
\begin{equation}
\mathcal{J}_{\text{GRPO}}(\theta) =
\mathbb{E}\!\left[
\frac{1}{G}\sum_{i=1}^{G} \hat{A}_i
\;-\; \beta\,\mathbb{D}_{\mathrm{KL}}\!\left(\pi_\theta \,\|\, \pi_{\text{ref}}\right)
\right],
\end{equation}
where $\hat{A}_i$ is the group-relative advantage of completion $i$ and $\beta$ weights
the KL term. We take $\pi_{\text{ref}}$ to be the base model with the LoRA adapter
disabled, which anchors the steered policy to the underlying model's language competence.
 
\paragraph{Optimisation and implementation.}
GRPO continues the Stage~1 LoRA adapter as a trainable policy; the base weights remain
frozen and only the adapter is updated. Completions are generated online with vLLM
\citep{vllm} in a co-located configuration that shares the GPU between
generation and training, and are sampled at temperature $1.0$ to ensure the in-group
diversity that GRPO learns from. Training is run on one
NVIDIA H100 GPU in bfloat16 with gradient checkpointing. Key hyperparameters are
listed in Table~\ref{tab:grpo-hparams}.
 
\begin{table}[t]
\centering
\small
\begin{tabular}{@{}lp{0.30\columnwidth}@{}}
\hline
\textbf{Hyperparameter} & \textbf{Value} \\
\hline
Base model & Hunyuan-MT-7B \\
Policy & continued Stage-1 LoRA adapter \\
LoRA rank / $\alpha$ / dropout & $16 / 32 / 0.05$ \\
LoRA target modules & q\_, k\_, v\_, o\_, gate\_, up\_, down\_proj. \\
\hline
Group size $G$ & 6 \\
Sampling temperature & 1.0 \\
Top-$p$ & 1.0 \\
KL coefficient $\beta$ & 0.04 \\
Clip range $\epsilon$ & 0.2 \\
\hline
Learning rate & $1\times10^{-6}$ \\
LR schedule & cosine, warmup 0.03 \\
Per-device batch size & 8 \\
Gradient accumulation & 4 \\
Generations reused / rollout & 2 \\
Training steps & 500 \\
Precision & bfloat16 \\
\hline
\end{tabular}
\caption{GRPO training configuration.}
\label{tab:grpo-hparams}
\end{table}

\subsubsection{Evaluation}
At inference time, all our different adapters are plugged into the unmodified RASTA pipeline of \citet{havaldar2025atowards}: source texts are embedded, the cultural alignment vector is applied, exemplars are retrieved, and the translation prompt is constructed exactly as in the original framework. Style alignment $\mathcal{A}$ is computed on the held-out test split using the same style quantifiers as in Section~\ref{3.1.1}, allowing direct comparison with the
published RASTA results for GPT-4. Because the GRPO reward is itself built from a
style quantifier of the same model family and training data as this evaluation
metric (see above), $\mathcal{A}$ gains for the GRPO variant cannot be treated as
fully independent evidence of style improvement: they show that the policy has
moved in the direction the automatic quantifier rewards, which we would expect to
correlate with, but not guarantee, a style shift a human rater would recognise. The
Base and SFT variants are not optimised against this quantifier and are not subject
to the same concern. To check that gains in style alignment do not come at the cost
of translation quality, we additionally report CometKiwi.


\section{Results}
\label{sec:results}
 
We evaluate steerability with the style alignment score $\mathcal{A}$ of
\citet{havaldar2025atowards}, i.e.\ the product--moment (Pearson) correlation between the
intended style of a source text and the realised style of its translation, as
measured by per-language style quantifiers. A score of $1$ indicates that
requested and realised style co-vary perfectly across the test set, and $0$
indicates no alignment. We report $\mathcal{A}$ for three variants of the
open-weight \mbox{Hunyuan-MT-7B} model: the unmodified \textbf{Base} model, the
supervised LoRA fine-tune (\textbf{SFT}), and the GRPO-trained model
(\textbf{GRPO}). Evaluation covers three styles (formality, politeness, and
intimacy) over twelve directed language pairs drawn from the same corpora used by
\citet{havaldar2025atowards}. To position an open $7$B model against a frontier system, we
additionally report the corresponding per-pair scores for vanilla GPT-4 and for
RASTA-augmented GPT-4, taken from \citet{havaldar2025atowards}.\footnote{GPT-4 values are
read from the per-pair heatmaps of \citet{havaldar2025atowards} (vanilla: Figures~4 and
A1; RASTA: Figure~A2), matched to the source$\to$target direction of each pair we
evaluate. Our evaluation uses the same metric $\mathcal{A}$, the same style
datasets, and the same train/test split as \citet{havaldar2025atowards}, and likewise
produces a single translation per source text. The comparison is therefore
like-for-like, and differences in $\mathcal{A}$ reflect the model alone.}
 
\subsection{Effect of fine-tuning}
\label{sec:results-finetuning}
 
Table~\ref{tab:finetuning} reports mean style alignment per style for each variant.
No variant dominates across the board. GRPO yields the only clear per-style gain,
raising formality alignment from $0.394$ (Base) to $0.473$, but it is neutral to
mildly negative on politeness ($0.498 \to 0.446$) and intimacy
($0.547 \to 0.528$); SFT tracks the Base model closely on every style. We flag,
however, that GRPO is trained with a reward derived from a style quantifier of the
same family used to compute $\mathcal{A}$ (Section~\ref{stage 2}), so this
formality gain is best read as evidence that GRPO successfully optimises against
the automatic style metric, not as an independently verified improvement in
style as a human rater would judge it; Base and SFT, which are not trained
against this signal, are not subject to this concern. Aggregated over all twelve
pairs, the three variants are statistically indistinguishable (Base $0.480$, SFT
$0.469$, GRPO $0.482$); a one-way ANOVA over the pair-level scores finds no
significant difference between methods ($F = 0.035$, $p = 0.966$).
 
The per-pair scores in Table~\ref{tab:comparison} show that this flat aggregate
hides, direction-dependent variation. Fine-tuning helps on
some pairs, formality en$\to$fr improves from $0.332$ (Base) to $0.548$ (GRPO),
and politeness en$\to$zh from $0.561$ (Base) to $0.659$ (SFT), but regresses
sharply on others, most notably politeness es$\to$ja, where alignment falls from
$0.374$ (Base) to $0.230$ (GRPO) and $0.210$ (SFT). Because the best variant
differs by pair, we also report a best-of selection (the highest-scoring
variant for each pair) as an upper bound on what per-pair model selection could
recover. This raises mean alignment to $0.522$ for politeness and $0.562$ for
intimacy (Table~\ref{tab:finetuning}), and to $0.519$ overall. 

\begin{table}[t]
\centering
\small
\begin{tabular}{lcccc}
\toprule
Style & Base & SFT & GRPO & Best-of \\
\midrule
Formality   & 0.394 & 0.415 & \textbf{0.473} & 0.473 \\
Politeness  & \textbf{0.498} & 0.453 & 0.446 & 0.522 \\
Intimacy    & \textbf{0.547} & 0.539 & 0.528 & 0.562 \\
\midrule
Overall     & 0.480 & 0.469 & \textbf{0.482} & 0.519 \\
\bottomrule
\end{tabular}
\caption{Mean style alignment $\mathcal{A}$ for the three Hunyuan-MT-7B variants,
averaged over the four directed language pairs per style. \emph{Best-of} selects
the highest-scoring variant per pair. Bold marks the best
single variant per row.}
\label{tab:finetuning}
\end{table}
 
\subsection{Open-weight versus GPT-4 under RASTA}
\label{sec:results-vs-gpt4}
 
Table~\ref{tab:comparison} compares the three Hunyuan-MT-7B variants against
RASTA-GPT-4 on every pair. With the RASTA method held fixed, the comparison
measures how much of the framework's benefit depends on the backbone model.
 
RASTA-GPT-4 is the stronger system overall, achieving the highest score on ten of
the twelve pairs and leading on every style average ($0.738$, $0.675$, $0.605$;
overall $0.672$ vs.\ a best Hunyuan variant of $0.482$ and a per-pair best-of of
$0.519$). The gap is widest on formality, where RASTA-GPT-4 ($0.738$) far exceeds
the best open variant ($0.473$), indicating that the open $7$B backbone captures
comparatively little of RASTA's formality signal.
 
The picture is closer on intimacy and on en$\leftrightarrow$zh directions, where
the open model is competitive and occasionally better. On both intimacy es$\to$fr
and fr$\to$es the open model surpasses RASTA-GPT-4 (Base $0.657$ vs.\ $0.650$; SFT
$0.622$ vs.\ $0.610$), and it comes within $0.03$ on politeness en$\to$zh
(SFT $0.659$ vs.\ $0.690$) and on intimacy zh$\to$en (Base $0.630$ vs.\ $0.670$).
The intimacy style average reflects this: RASTA-GPT-4 leads by
only $0.043$ over the per-pair best-of open model ($0.605$ vs.\ $0.562$), the
narrowest margin of the three styles.
 
\begin{table}[t]
\centering
\small
\setlength{\tabcolsep}{5pt}
\begin{tabular}{lcccc}
\toprule
& \multicolumn{3}{c}{Hunyuan-MT-7B} & GPT-4 \\
\cmidrule(lr){2-4}\cmidrule(lr){5-5}
Pair & Base & SFT & GRPO & RASTA \\
\midrule
\multicolumn{5}{l}{\emph{Formality}} \\
\quad en$\to$fr & 0.332 & 0.485 & \underline{0.548} & \textbf{0.770} \\
\quad fr$\to$en & 0.406 & 0.428 & \underline{0.454} & \textbf{0.780} \\
\quad it$\to$pt & 0.444 & 0.419 & \underline{0.488} & \textbf{0.780} \\
\quad pt$\to$it & 0.395 & 0.329 & \underline{0.402} & \textbf{0.620} \\
\quad \textit{Avg.} & 0.394 & 0.415 & 0.473 & 0.738 \\
\addlinespace
\multicolumn{5}{l}{\emph{Politeness}} \\
\quad en$\to$zh & 0.561 & \underline{0.659} & 0.637 & \textbf{0.690} \\
\quad zh$\to$en & \underline{0.560} & 0.531 & 0.542 & \textbf{0.720} \\
\quad es$\to$ja & \underline{0.374} & 0.210 & 0.230 & \textbf{0.630} \\
\quad ja$\to$es & \underline{0.496} & 0.412 & 0.375 & \textbf{0.660} \\
\quad \textit{Avg.} & 0.498 & 0.453 & 0.446 & 0.675 \\
\addlinespace
\multicolumn{5}{l}{\emph{Intimacy}} \\
\quad zh$\to$en & \underline{0.630} & 0.577 & 0.611 & \textbf{0.670} \\
\quad en$\to$zh & 0.286 & \underline{0.338} & 0.308 & \textbf{0.490} \\
\quad es$\to$fr & \underline{\textbf{0.657}} & 0.619 & 0.578 & 0.650 \\
\quad fr$\to$es & 0.616 & \underline{\textbf{0.622}} & 0.616 & 0.610 \\
\quad \textit{Avg.} & 0.547 & 0.539 & 0.528 & 0.605 \\
\midrule
\textbf{Overall} & 0.480 & 0.469 & 0.482 & 0.672 \\
\bottomrule
\end{tabular}
\caption{Per-pair style alignment $\mathcal{A}$ for the three Hunyuan-MT-7B
variants against RASTA-GPT-4. All systems use the RASTA prompt; GPT-4 values are
from \citet{havaldar2025atowards}. \textbf{Bold} marks the best system in each row;
\underline{underline} marks the best Hunyuan variant in each row. \emph{Avg.}\ and
\emph{Overall} are unweighted means over pairs.}
\label{tab:comparison}
\end{table}

\subsection{Translation quality}
\label{sec:results-quality}
 
Style alignment is only useful if the underlying translation remains adequate, so
we also measure COMETKIWI (CK). Table~\ref{tab:quality} reports per-pair CK with
per-style means. The result is unambiguous: the Base model attains the highest CK on
all twelve pairs, and both fine-tuned variants degrade quality consistently.
Averaged over all pairs, CK falls from $0.777$ (Base) to $0.689$ (GRPO) and $0.681$
(SFT), a drop of roughly $0.09$--$0.11$ on formality and politeness and $0.08$ on
intimacy. The largest single degradation is on politeness ja$\to$es, where CK falls
from $0.768$ (Base) to $0.602$ (GRPO).

For reference, \citet{havaldar2025atowards} report a CK score in the
range of $0.72$--$0.80$ across styles with GPT-4, so the Base model's quality ($0.78$ overall) is
broadly comparable. 
 
\begin{table}[t]
\centering
\small
\begin{tabular}{lccc}
\toprule
Pair & Base & SFT & GRPO \\
\midrule
\multicolumn{4}{l}{\emph{Formality}} \\
\quad en$\to$fr & \textbf{0.834} & 0.765 & 0.768 \\
\quad fr$\to$en & \textbf{0.817} & 0.741 & 0.735 \\
\quad it$\to$pt & \textbf{0.792} & 0.644 & 0.657 \\
\quad pt$\to$it & \textbf{0.781} & 0.647 & 0.679 \\
\quad \textit{Avg.} & \textbf{0.806} & 0.699 & 0.710 \\
\addlinespace
\multicolumn{4}{l}{\emph{Politeness}} \\
\quad en$\to$zh & \textbf{0.809} & 0.745 & 0.778 \\
\quad zh$\to$en & \textbf{0.760} & 0.726 & 0.698 \\
\quad es$\to$ja & \textbf{0.777} & 0.652 & 0.661 \\
\quad ja$\to$es & \textbf{0.768} & 0.594 & 0.602 \\
\quad \textit{Avg.} & \textbf{0.778} & 0.679 & 0.685 \\
\addlinespace
\multicolumn{4}{l}{\emph{Intimacy}} \\
\quad zh$\to$en & \textbf{0.747} & 0.681 & 0.684 \\
\quad en$\to$zh & \textbf{0.777} & 0.720 & 0.725 \\
\quad es$\to$fr & \textbf{0.727} & 0.625 & 0.640 \\
\quad fr$\to$es & \textbf{0.739} & 0.627 & 0.639 \\
\quad \textit{Avg.} & \textbf{0.748} & 0.663 & 0.672 \\
\addlinespace
\midrule
\textbf{Overall} & \textbf{0.777} & 0.681 & 0.689 \\
\bottomrule
\end{tabular}
\caption{Per-pair COMETKIWI translation quality for the three Hunyuan-MT-7B variants
(all under RASTA). The Base model is best on every style and on all twelve
individual pairs; \textbf{bold} marks the best variant per row. \emph{Avg.}\ and
\emph{Overall} are unweighted means over pairs.}
\label{tab:quality}
\end{table}



\section{Discussion}
\label{sec:discussion}

Our research question asks whether a fine-tuned open-weight model, specifically
Hunyuan-MT-7B, can achieve style alignment comparable to GPT-4 when integrated
into the RASTA framework. The results give a qualified answer. Aggregated over all
twelve language pairs and three styles, RASTA-GPT-4 remains the stronger system
($\mathcal{A} = 0.672$) than any open-weight variant (best single variant $0.482$;
per-pair best-of $0.519$). However, this overall gap is not uniform: it is large
for formality, modest for politeness, and small for intimacy, where the open model
is competitive and on the Spanish--French pairs exceeds RASTA-GPT-4 outright.
Comparable alignment is therefore achievable for some styles and language
directions, but not as a blanket replacement for GPT-4 across the board.
Nevertheless, the fact that a 7B-parameter open-weight model can approach GPT-4's performance 
in several settings suggests that smaller, substantially more affordable models are a 
promising direction for style-controlled machine translation within the RASTA framework.

\subsection{Fine-tuning did not reliably improve style alignment}
\label{sec:disc-finetuning}

The most direct finding is that neither supervised fine-tuning nor GRPO produced a
consistent improvement over the base model. The three variants are statistically
indistinguishable in aggregate ($F = 0.035$, $p = 0.966$), and the only clear
per-style gain, GRPO on formality, is offset by neutral-to-negative effects
on politeness and intimacy. At the pair level the effect is strongly
direction-dependent: GRPO lifts formality en$\to$fr from $0.332$ to $0.548$, yet
fine-tuning sharply degrades politeness es$\to$ja, from $0.374$ (Base) to $0.230$
(GRPO) and $0.210$ (SFT).

Several factors plausibly contribute. First, because all variants are evaluated
inside the RASTA scaffold, the retrieved exemplars and style conditioning already
supply much of the steering signal; weight updates may add little on top of a
prompt that is itself doing the heavy lifting. Second, the GRPO reward is derived
from the same style quantifiers used for evaluation, so noise or miscalibration in
a quantifier, can push the policy in unhelpful directions. 

\subsection{The gap is concentrated in formality}
\label{sec:disc-gap}

Holding RASTA fixed, Hunyuan-MT-7B trails GPT-4 most on formality
($0.473$ vs.\ $0.738$) and least on intimacy ($0.562$ vs.\ $0.605$ for the best-of
open model). Two features of the formality setting likely amplify this gap.
Formality is the noisiest of the three constructs to measure: \citet{havaldar2025atowards}
report the highest style-quantifier RMSE for formality, and its labels are binary
rather than continuous, making the correlation-based $\mathcal{A}$ a coarser signal
than for politeness or intimacy.

\subsection{Limitations}
\label{sec:disc-limitations}

Our evaluation reuses the style quantifiers of \citet{havaldar2025atowards}, which are
imperfect regressors and serve as both reward and metric, coupling their errors to
our results. A further limitation concerns our model. Hunyuan-MT-7B is a
translation specialist, optimised to produce faithful, register-neutral
renderings across its supported languages. This specialisation is a double-edged
property in our setting. On one hand it gives strong underlying translation quality:
Table~\ref{tab:quality}). On the other hand, a model whose training objective rewards
semantic fidelity rather than stylistic variation may carry a strong prior toward a
single ``default'' translation, and this prior could weaken its responsiveness to the
style conditioning that RASTA and our GRPO reward attempt to induce. A
general-purpose instruction-following model such as GPT-4 has broader latitude to
reshape its output in response to in-context exemplars and an explicit style level.
We cannot separate, from our experiments alone, how much of the model gap
reflects raw model scale and how much reflects this specialisation, because the two
are confounded: our open model is both substantially smaller than GPT-4 and a
narrow translation specialist.

A second limitation is the scale of the GRPO stage. We ran GRPO for only a modest
number of updates (Table~\ref{tab:grpo-hparams}), with a small learning rate
($1\times10^{-6}$) and a KL penalty anchoring the policy to the frozen reference. Under
these settings the policy can move only a short distance from its supervised
initialisation. The near-identical aggregate scores of the SFT and GRPO variants are therefore
consistent with under-training rather than with a genuine ceiling on what
GRPO can achieve for this task.

These limitations point to concrete next steps. First, the GRPO stage should be
scaled substantially, with more training steps to test directly whether the flat
SFT/GRPO result is an artefact of under-training rather than a property of the
method. Second, and more informatively, the pipeline should be repeated with a
general-purpose open instruction-tuned model of comparable size (for example a
Qwen or Llama variant in the 7-8B range) in place of the translation specialist.
Comparing a specialist and a generalist at the same parameter count would isolate
whether the limited steerability we observe stems from translation specialisation
or simply from scale, and would clarify which kind of open model is the more
promising substitute for GPT-4 within RASTA. Third, because our style alignment
metric $\mathcal{A}$ and our GRPO reward both rely on the same family of
automatic style quantifiers (Section~\ref{sec:disc-limitations}), a small-scale
human evaluation is needed to confirm that the gains we report reflect
style preservation a person would actually notice. We should check the
pairs where the open model matches or exceeds RASTA-GPT-4 under $\mathcal{A}$,
namely intimacy es$\to$fr and fr$\to$es and the en$\leftrightarrow$zh politeness
and intimacy directions (Section~\ref{sec:disc-gap}), since these are the cases
where our central claim, that an open model can be competitive with GPT-4, most
needs independent, human-judged confirmation.

\section{Conclusion}
\label{sec:conclusion}

We investigated whether an open-weight, translation-specialised model can substitute
for GPT-4 as the generator inside RASTA, the retrieval-augmented stylistic alignment
framework of \citet{havaldar2025atowards}. Using Hunyuan-MT-7B and evaluating three
variants: base, LoRA-SFT, and GRPO, across formality, politeness, and
intimacy, we reach two main conclusions.

First, targeted fine-tuning of the backbone did not reliably improve style
alignment: the base, SFT, and GRPO variants are statistically indistinguishable in
aggregate, with gains on some pairs cancelled by regressions on others.

Second, an open $7$B backbone within RASTA does not match GPT-4's overall style
alignment, but the shortfall is style- and direction-specific rather than uniform:
it is large for formality, where measurement is noisiest and the register is
hardest to express, and small for intimacy and Chinese directions, where the open
model matches or surpasses RASTA-GPT-4 on individual pairs. Because neither
fine-tuned variant improved style alignment reliably, and both consistently
degraded CometKiwi relative to the unmodified Base model, our results point to the
Base model as the most practically useful of the three open-weight variants: it
keeps translation quality highest while still achieving style alignment
comparable to the fine-tuned alternatives. Because our alignment metric and our GRPO reward draw on the same family of automatic style quantifiers,
we see a small, targeted human evaluation, focused
on the pairs where the open model already matches or beats RASTA-GPT-4, as an important next step for establishing whether these automatic gains
correspond to style preservation a human reader would actually perceive.

\bibliography{custom}

\appendix

\section{Example Appendix}
\label{sec:appendix}

\subsection{Data used from OPUS}
\label{sec:dataused}

For every language pair we used 20000 sentence pairs from the TED2020 dataset \cite{4992de1b5fb34f3e9691772606b36edf}. For en-zh this contained a mix of different sorts of Chinese, Mandarin and Cantonese. For this pair we used the same amount of rows from UNPC data which is only Mandarin.

\subsection{Translation Prompt}\label{prompt}
This is the prompt used in LoRA finetuning stage \ref{stage 1}:

\noindent
Your task is to translate a given piece of text from the source language (\texttt{src}) to the target language (\texttt{tgt}).

Keep in mind that style and tone vary across cultures, so a direct word-for-word translation may not always preserve the intended meaning and register of the original text.

The text to be translated is:

\begin{quote}
\texttt{src\_text}
\end{quote}

Translate the text into the target language (\texttt{tgt}). Output only the translation.\

\paragraph{Github}
The code can be found on my Github: https://github.com/ThomasR2004/MasterThesis
\end{document}